\documentclass[a4paper,12pt]{report}

\usepackage{latexsym}
\usepackage{graphicx}
\usepackage{makeidx}
\usepackage{amsmath}
\usepackage{epsfig}
\usepackage{geometry}
\title{
\vspace{-4.5cm}
\textbf{Semi-Automatic Segmentation and Ultrasonic Characterization of Solid Breast Lesions} \vspace{0.5cm}}
\author{
{Mohammad Saad Billah (0706063)} \\
{Tahmida Binte Mahmud (0706061)}
\vspace{2.5cm}\\
\small{A thesis submitted to the Department of Electrical and Electronic Engineering}\\
\small{of}\\
\small{Bangladesh University of Engineering and Technology}\\
\small{in partial fulfillment of the requirement for the degree of} \\
\vspace{0.01cm}\\
\small{BACHELOR OF SCIENCE IN ELECTRICAL AND ELECTRONIC ENGINEERING}\\
\vspace{1cm}
{ Submitted to}\\
{Dr. Md. Kamrul Hasan}\\
{Professor}
\vspace{1.5cm}\\
\small{DEPARTMENT OF ELECTRICAL AND ELECTRONIC ENGINEERING}\\
\vspace{0.01cm}\\
\vspace{0.5cm}
\small{ BANGLADESH UNIVERSITY OF ENGINEERING AND TECHNOLOGY  } \\
\small{ Dhaka 1000, Bangladesh} \\ %\vspace{1cm} \\
{ February, 2013} %\vspace{1cm} \\
}
\date{}

\begin{document}
%\pagestyle{empty}
%\input Title.tex
%
%\newpage
%\pagestyle{plain}
\maketitle

\pagenumbering{roman}
%\setcounter{page}{1}
%\setcounter{secnumdepth}{-1}
%\input{Declaration.tex}
%\normalsize\pagenumbering{roman}
%\newpage
%\begin{center}
%\input{Dedication.tex}
%\end{center}
%\newpage

%\input{Acknowledgements.tex}

\newpage
\tableofcontents
\newpage
\listoffigures
\newpage
\listoftables
\newpage

\begin{center}
\section{Abstract}
\end{center}
Characterization of breast lesions is an essential prerequisite to detect breast cancer in an early stage. Automatic segmentation makes this categorization method robust by freeing it from subjectivity and human error. Both spectral and morphometric features are successfully used for differentiating between benign and malignant breast lesions. In this thesis, we used empirical mode decomposition method for semi-automatic segmentation. Sonographic features like ehcogenicity, heterogeneity, FNPA, margin definition, Hurst coefficient, compactness, roundness, aspect ratio, convexity, solidity, form factor were calculated to be used as our characterization parameters. All of these parameters did not give desired comparative results. But some of them namely echogenicity, heterogeneity, margin definition, aspect ratio and convexity gave good results and were used for characterization.

\newpage
\normalsize \pagenumbering{arabic}
%\mainmatter
\setcounter{secnumdepth}{2}

\chapter{Introduction}
Breast cancer is a major threat to female health all over the world. The mortality rate can be significantly reduced if cancer lesions could be detected and treated in no time. This requires a regular screening mechanism of adult women via a noninvasive, hazard-free, low-cost and available as well as reliable medical imaging modality. Until recently, mammographic evaluation was being considered as the noninvasive gold standard for breast mass diagnosis. However, as ultrasonography is pain and radiation-hazard free, and because of the advent of very high resolution ultrasound equipment, it is increasingly becoming popular for breast mass diagnosis. Moreover, sonographic evaluation is also known to be superior for dense breast lesion detection.

\section{Significance of the Thesis}
\begin{figure}[!h]
  \centering
  \includegraphics[width=4in,angle=0]{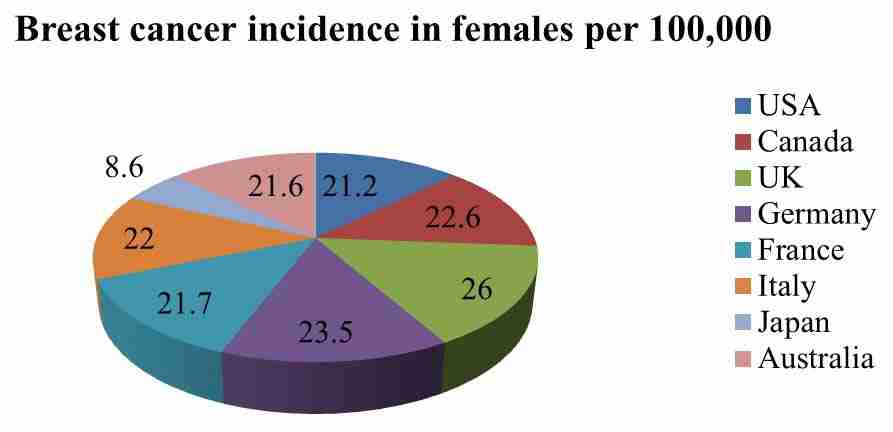}\\
  \caption{Global scenario of brest cancer.}\label{Fig:pie1}
\end{figure}

From a study of 2010, nearly 1.5 million people were affected globally by breast cancer \cite{wwcancer}. It is the principle cause of death from cancer among women all over the world. However one-third of these deaths could be decreased if detected and treated early \cite{wwcancer}. In 2009, 48,417 women and 371 men in the UK were diagnosed with breast cancer \cite{btype}. 11,556 women and 77 men in the UK died from breast cancer in 2010 \cite{btype}. Alarmingly high 76,000 women died from breast cancer in South Asia \cite{lstory}. About 1 in 8 U.S. women (just under $12\%$) will develop invasive breast cancer over the course of her lifetime \cite{stat}. A global scenario is shown in figure \ref{Fig:pie1} \cite{inci}.

In Bangladesh, the rate of occurrence of breast cancer is increasing at an alarming rate because of the adoption of western lifestyle such as higher fat diets, reduced activity, delayed marriage and child bearing, and decreased breast feeding. Sixteen percent of the total cancer affected women in the country are victim to breast cancer, says a World Health Organization (WHO) study. About $35,000$ women in Bangladesh develop breast cancer every year \cite{us_em}. About 15,000 breast cancer patients die every year in Bangladesh \cite{fin_ex}. Based on the data available from the Radiotherapy Department of the Dhaka Medical College Hospital, it is estimated that the incidence of breast cancer will be about $17\%$ of the total cancer patients. WHO also ranked Bangladesh 2nd in terms of mortality rate of women in the countries suffering from breast cancer.

\section{Objectives of the Thesis}
Saving lives of several hundreds of thousands of women suffering from breast carcinoma around the world is only possible if tumors can be diagnosed at an early stage. The objectives of this thesis, therefore, are

\begin{enumerate}
\item To improve the current state of breast cancer diagnosis around the globe by detecting cancerous tissues at an early stage using ultrasonic features.
\item To develop a semi-automatic segmentation algorithm for ensuring robust extraction of ultrasonic features.

\end{enumerate}

\section{Physics of Ultrasound}
Modern ultrasound imaging started its journey from World War II Navy sonar technology. Ultrasound technology advanced through the 1960s from simple A-mode and Bmode scans to today's M-mode and Doppler two-dimensional and three-dimensional systems. Ultrasound refers to sound with frequencies higher than the highest audible frequency for human beings. So, any sound above about 20KHz is considered to be ultrasound. But medical ultrasound systems typically operate between 1 and 10 MHz. The principles of ultrasound propagation are similar to those of ordinary sound propagation and are defined by the theory of acoustics. Ultrasound moves like a wave by expansion and compression of the medium. Ultrasound waves travel at a certain speed, depending on the the traveling material. These waves can be absorbed, refracted, focused, reflected, and scattered. A transducer converts electrical signals to acoustic signals. It generates pulses of ultrasound and sent through a patient's body. Organ boundaries and complex tissues produce echoes by reflection or scattering. The echoes return back and get detected by the transducer. Then the acoustic signal is converted to an electrical signal. The echoes are then processed by the ultrasound imaging system and a grayscale image of human anatomy is produced on a display. Each point in the image corresponds to the echo strength.

A succession of these signals can be displayed on an oscilloscope by repetitive firing of the transducer. This display is called the A-mode scan. A B-mode scan is generated by scanning the transducer beam in a plane. The transducer is moved in x-direction while its beam is aimed down the z-axis. The dominant B-mode imaging method in early ultrasound imaging was scanning a single transducer. However, three types of B-mode scanners presently dominate namely linear scanners, mechanical sector scanners and phased array sector scanners \cite{prince}.

\section{Ultrasound Image Segmentation: Background and Literature}
Ultrasound images sometimes display poor quality because of multiplicative speckle noise that results in artifacts. Segmentation of lesions in ultrasound images is a research area where desired accuracy is yet to be achieved. In regular breast screening approaches, the suspected region is manually located by a trained radiologist. A rectangular region of interest (ROI) is then chosen by the radiologist. A computer-aided diagnosis (CAD) system is used for further analysis leading to the classification of the tumor. Inaccurate selections of ROI can severely affect the performance of the CAD system. If this selection includes such level of human involvement, these steps are open to subjectivity and human error. It is therefore a challenging task to provide the radiologist with an automated tool that can effectively assist in the selection of the ROI to improve the consistency of diagnosis. This automatic detection of ROIs is not intended to replace the radiologist, but rather to detect the ROIs efficiently in less time which might otherwise be missed if located with human eye. For automatic segmentation and lesion extraction several techniques have been attempted so far. Drukker et al. used the radial gradient index (RGI) filtering technique to automatically detect lesions from breast ultrasound images and with an overlap level of 0.4 with lesions outlined by a radiologist, $75\%$ accuracy of lesion detection was achieved \cite{drukker}. Yap et al. analyzed the use of statistical methods and values of fractal dimensions. The images were preprocessed using histogram equalization; hybrid filtering and marker-controlled watershed segmentation were applied \cite{yap}. The accuracy of ROI detection when using local mean was $69.21\%$ and $54.21\%$ using fractal dimension. Chang et al. used watershed segmentation algorithm for automatic segmentation \cite{chen}. In our thesis, a novel approach to initial lesion detection in ultrasound breast images is proposed. The novelty of our approach lies in the use of empirical mode decomposition (EMD) which had never been used before in automatic segmentation of lesions and ROI extraction. Histogram equalization has been used in the preprocessing stage, followed by diffusion filtering, empirical mode decomposition, automatic thresholding using intraclass variance minimization method and boundary drawing approach for ROI labeling.

\section{Tissue Characterization: Background and Literature}
Breast ultrasound has been successfully characterizing breast lesions by categorizing them into a number of distinctive groups based on their benign and malignant features. This categorization demonstrates the relative risk for malignancy and determines the need for biopsy. Several ultrasound features have the potential to be used as lesion characterization parameters. Among the spectral features echogenicity, heterogeneity, FNPA, cooccurence contrast, Hurst coefficient, margin definition have been used and among the morphometric features aspect ratio, lesion area, compactness, roundness, convexity, solidity and form factor are popular \cite{stravos}. Kobayashi et al. (1979) and Harper et al. (1982) used acoustic shadowing as their characterization feature. Drukker et al implemented shadowing as the feature and obtained $80\%$ sensitivity \cite{drukker2}. Joo, Lefebvre and Alam obtained an area of 0.95, 0.85 and 0.947 respectively under ROC curve based on spiculation, branch patterns and number of lobulations \cite{joo}, texture and morphometric parameters \cite{lefebv} and a combination of heterogeneity, convexity, margin definition and fractal dimension \cite{kalam}. Stavros et al \cite{stravos2}, Skaane et al \cite{skaane}, Huber et al \cite{huber}, Garra et al \cite{garra} have worked with different sonographic features for breast tissue characterization as well.

\section{Organization of the Thesis}

This thesis consists of five chapters. Chapter 1 gives a brief description about the significance of our thesis, the objectives of the thesis, the physics of ultrasound and the background and literature of ultrasound image segmentation and tissue characterization. Chapter 2 includes our semi-automatic segmentation algorithm using empirical mode decomposition. In Chapter 3, we present the detailed discussion about the extraction algorithm of ultrasonic (both spectral and morphometric) features. Chapter 4 includes the data acquisition process, the result of semi-automatic segmentation process, the comparative values of ultrasonic features for benign and malignant tissues. It also presents the comparative sensitivity, specificity and receiver operating characteristics of ultrasonic features. Finally, the chapter includes correlation with histopathologic findings. Chapter 5 concludes the thesis by presenting the overall view of the thesis and pointing out our limitations and some scope for future improvements.

\newpage
\chapter{Semi-Automatic Segmentation Using Empirical Mode Decomposition}
For robust characterization of breast tissues, it is essential to locate the suspected region accurately. Inaccurate selection of ROI due to subjectivity or human error can degrade the performance of the characterization algorithm. We, therefore, tried to develop a semi-automatic method of segmentation so that the characterization process become as accurately as possible. We have used an empirical mode decomposition (EMD) based technique for segmentation.

\section{Materials and Methods}
%The database used in our segmentation process comprises of 100 cases including both benign and malignant ones. A SonixTOUCH Research ultrasound machine with L14-5/38 linear transducer has been used in scanning patients. Two expert sonologists actively participated in the data collection process. Each image had been manually processed by the sonologists and the extreme points of the suspected lesions were already marked.

The complexity of ultrasound images lies in data composition, which is described in terms of speckle information. Speckle noise consists of a relatively high-level gray intensity, qualitatively ranging between hyperechoic (bright) and hypoechoic (dark) domains \cite{friedland}. So, the images need to be pre and post-processed to reduce the speckle noise and increase the dynamic range.

\begin{figure}[!h]
  \centering
  \includegraphics[width=4in,angle=0]{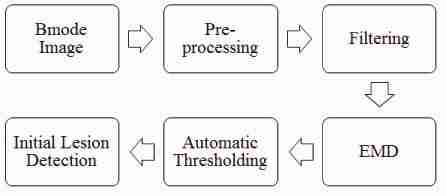}\\
  \caption{Modular block diagram of proposed method.}\label{Fig:block}
\end{figure}

Figure \ref{Fig:block} shows a modular block diagram of our proposed technique.

\subsection{Preprocessing}
The experience of the examiner and the quality of the original image i.e. the scanner are the two most important factors upon which the credibility of the scanning is largely depended. Homogeneity of the original ultrasound image is an essential prerequisite for accurate lesion ROI detection. This is accomplished in the preprocessing stage. We have used a histogram equalization strategy tested in earlier experiments \cite{wintz} as a preprocessing stage. Histogram equalization attempts to increase the dynamic range of the pixel values in an image as contrast stretching does. However, unlike contrast stretching, histogram equalization does not involve interactivity as same result is produced when it is applied to an image with a fixed number of bins.

Let us consider a discrete grayscale b-mode image $X$ and let $n_l$ be the number of occurrences of gray level $l$. The probability of an occurrence of a pixel $(i,j)$ of level $l$ in the image is

\begin{equation}
   p_{X_{ij}}(l)= p(X(i,j)=l) = \frac{n_l}{N},0\leq l<L   \label{hist} \nonumber
\end{equation}

\noindent $L$ being the total number of gray levels in the image, $N$ being the total number of pixels in the image, and $p_{X_{ij}}(l)$ being in fact the image's histogram for pixel value $l$, normalized to $[0,1]$. Let us also define the cumulative distribution function corresponding to $p_{X_{ij}}$ as

\begin{equation}
   F_{X_{ij}}(l)= \sum_{m=0}^{l} p_{X_{ij}}(m)  \label{hist2} \nonumber
\end{equation}

\noindent which is also the image's accumulated normalized histogram. A transformation of the form $Y = T(X)$ can be created to produce a new image $Y$, such that its CDF will be linearized across the value range, i.e.

\begin{equation}
   F_{Y_{ij}}(l)= lK  \label{hist3}    \nonumber
\end{equation}

\noindent for some constant K. The properties of the CDF allows to perform such a transform; it is defined as

\begin{equation}
   Y = T(X) = F_{X}(X)  \label{hist4}  \nonumber
\end{equation}

\noindent Here it is noticed that the T maps the levels into the range [0,1]. In order to map the values back into their original range, the following simple transformation needs to be applied on the result:

\begin{equation}
  Y_h = Y (max(max(X)) - min(min(X))) + min(min(X))
\end{equation}

\begin{figure}[!h]
  \centering
  \includegraphics[width=5in,angle=0]{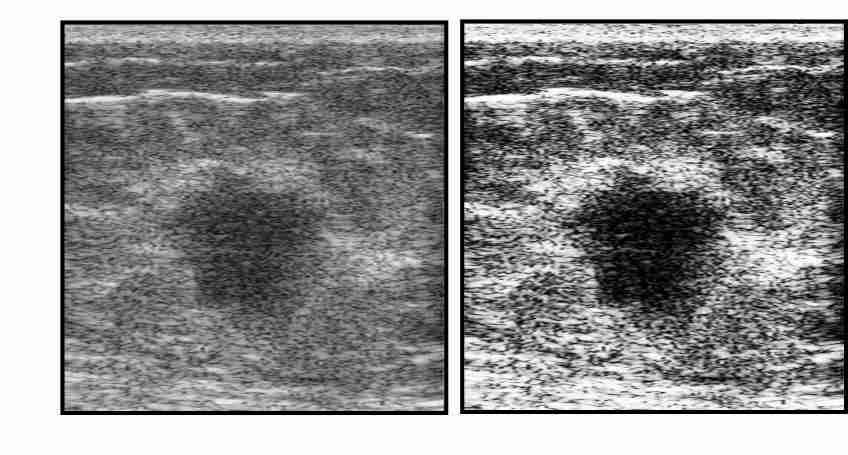}\\
  \caption{Original bmode image and image after histogram equalization.}\label{Fig:1}
\end{figure}

%
%\begin{figure}[!h]
%  % Requires \usepackage{graphicx}
%  \centering
%  \includegraphics[width=2.2in,angle=0]{histeq.eps}\\
%  \caption{Bmode image after histogram equalization.}\label{Fig:2}
%\end{figure}

%
%
%\begin{equation}
%   p_{r}(r_{j}),j = 1,2,3\cdots L    \label{hist}
%\end{equation}
%
%
%Let \ref{hist} denote the histogram associated with the intensity levels of a given image, and recall that the values in a normalized histogram are approximations to the probability of occurrence of each intensity level in the image [ref].
%
%\begin{equation}
%   s_k = T(r_{k}) = \sum_{j=1}^{k} p_{r}(r_{j}) = \sum_{j=1}^{k} \frac{n_j}{n}  \label{hist2}
%\end{equation}
%
%For discrete quantities, the equalization transformation becomes \ref{hist2} for $k = 1,2,\cdots,L,$ where skis the intensity value in the output (processed) image corresponding to value $r_k$ in the input image, n is the total number of pixels, and $n_j$ is the number of pixels in bin j [ref].

\subsection{Diffusion filtering}
As noise is a major obstacle to accurate segmentation of the images, the removal of noise is a vital process ensured in the filtering stage. Median filtering has been a widely used approach for removing speckle noise in ultrasound images. However, according to Yap et al. \cite{yap2} the inaccuracy of the boundary detection by Drukker et al. \cite{drukker}, Joo et al. \cite{joo} and Kupinski et al. \cite{kupinski} partially depended on their use of median filters as  along with the speckle noise, the important edge information-in particular, edges that belonged to the lesion was also lost. Gaussian blur \cite{chen_dr} is a linear filtering technique that is popular to reduce the oversegmentation problem in ultrasound images. In spite of being effective in removing speckle noise, this algorithm blurs and dislocates edges \cite{mrazek} which may negatively affect subsequent lesion segmentation. Perona and Malik \cite{peronamalik} proposed a nonlinear partial differential equation approach for smoothing images on a continuous domain. Anisotropic diffusion tends to perform well for images corrupted by additive noise. In our  proposed algorithm, we used a geometric nonlinear diffusion filtering approach proposed by Gonzalez et al which rather than employing four directional gradients around the pixel of interest, uses geometric parameters derived from the local pixel intensity distribution in calculating the diffusion coefficients in the horizontal and vertical directions \cite{gonzo2}. The filter generates output $Y_{d}$ from input signal $Y_h$ after $n$ iterations as follows

\begin{eqnarray}
Y_{d} = Y^{n}_{h} &=& Y^{n-1}_{h} + \triangle k [c(D_x,P_x).(\nabla_E + \nabla_W)+ c(D_y,P_y).(\nabla_N + \nabla_S)]^{n-1}
\end{eqnarray}

\noindent Here,$Y^{0}_{h}$ is the original histogram equalized image and $Y^{n}_{h}(n > 0)$ is the diffused strain image at the $n$th step,$\nabla_p = I_p - I_s(p= E,W, N$ and $S )$ denotes the difference between the interrogative pixel and one of the east, west, north and south pixels, respectively, $n$ is the iteration number, $\triangle k$ is the integration constant and $C$ is the wregion diffusivity function defined as

\begin{eqnarray}
C(D_{a},P_{a}) = \frac{1}{1+(\frac{D_{a}}{|P_{a}| + \epsilon})^2}   \nonumber
\end{eqnarray}

\noindent where $a$ represents $x$ or $y$ directions, $D_a$ denotes the $a$-directional intensity difference in a $3 \times 3$ window and $P_a$ is defined from $D_a$ depending on a threshold value of the image intensity of the interrogative pixel. We use $1 \leq n \leq 15$ and $\triangle k = 0.2$.

%Subsection D, next, provides a brief overview of empirical mode decomposition and associated analysis methodology.

\subsection{Empirical mode decomposition}
EMD, developed by Huang et al. \cite{huang} relies on a fully data-driven mechanism excluding the need of any a priori known basis. It decomposes a signal into a sum of intrinsic mode functions (IMFs).

EMD is performed on 1-D signal which can be written in the form where $i$ represents the axial depth index and $j$ represents envelope A-line index.

\begin{equation}
r^{j}(i) = Y_{d}(i,j)    \mbox{ for } 1 \leq i \leq S_{i} \mbox{ and } j =j_{s}
\end{equation}

\noindent where $S_{i}$ represents the length of the 1-D  envelope line. Given a signal $r^{j}(i)$, for our case the diffusion filtered signal, first the extrema of $r^{j}(i)$ are detected. The upper and lower envelopes $u^{j}(i)$ and $l^{j}(i)$ are generated by connecting the maxima and minima separately with cubic spline interpolation. Then the local mean is determined as $m^{j}(i)=[u^{j}(i)+l^{j}(i)]/2.$ IMF should have zero local mean. So, $m^{j}(i)$ is subtracted from $r^{j}(i)$ to get the first component $r^{j}_{1}(i)=r^{j}(i)-m^{j}(i).$ To find rest of the IMF components, residue $res_1^{j}(i)$ is generated by subtracting $r^{j}_{1}(i)$ from signal $r^{j}(i)$ as $res_{1}^{j}(i)=r^{j}(i)- r^{j}_{1}(i).$ The sifting process is continued until the desired IMFs are extracted from the signal. At the end if the sifting process, the signal $r^{j}(i)$ can be represented as

\begin{equation}
r^{j}(i) = \sum_{q=1}^{g} r^{j}_{q}(i) + res^{j}_{g}(t)  \mbox{ for } 1 \leq i \leq S_{i}
\end{equation}

\begin{figure}[!h]
  \centering
  \includegraphics[width=5.5in,angle=0]{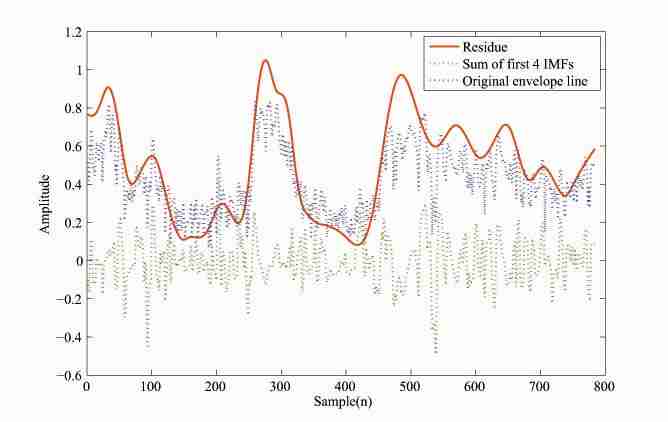}\\
  \caption{An RF line envelope, sum of its first 4 IMFs and the residue.}\label{imf_fig}
\end{figure}

\noindent It is illustrated in figure \ref{imf_fig} for one envelope of RF line after preprocessing and filtering. Lower order IMFs capture fast oscillation modes while higher order IMFs typically represents low oscillation modes. Figure \ref{imf_fig} also shows the typical first four IMFs of the EMD of one envelope data. While the lower order IMFs are high frequency component of the envelope thus capturing the speckle noise and texture, the higher order and lower frequency IMFs captures more slow variation like the base echogenicity of the tissue type behind the texture and noise information. Thus the higher order is able to approximate the boundaries and reduce speckles as well as improve the edge information in the ultrasound images at the cost of losing some texture information. In this method, the residue signal after taking out 4 IMFs is used in the next step of processing.

\begin{equation}
res^{j}_{4}(i) = r^{j}(i) - \sum_{q=1}^{4} r^{j}_{q}(i)   \mbox{ for } 1 \leq i \leq S_{i}
\end{equation}

\begin{figure}[!h]
  \centering
  \includegraphics[width=5.2in,angle=0]{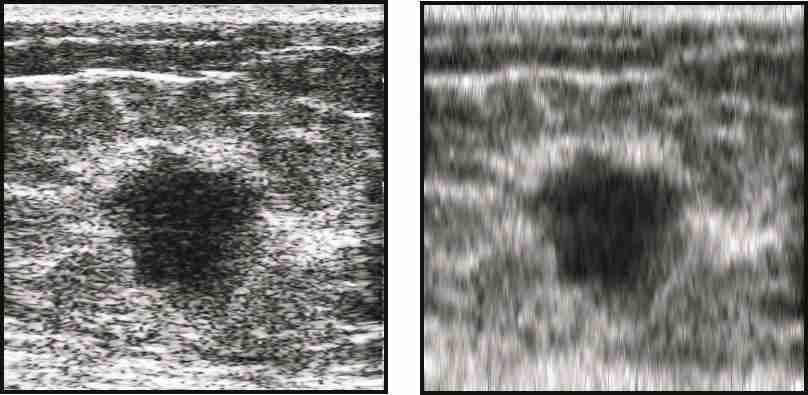}\\
  \caption{Diffusion filtered image before and after EMD.}\label{Fig:3}
\end{figure}

\subsection{Automatic thresholding}
An image is divided into its constituent parts through segmentation. Thresholding segmentation \cite{sahoo} is a popular algorithm known for its simplicity and time intensive nature. An intensity value called the "threshold" is determined which separates pixels into desirable classes. Within our present research context, to turn an ultrasonic image into a binary one in order to separate the tumor from its background, an automatic threshold-determination method, proposed by N. Otsu has been adopted, which can choose the threshold to minimize the intraclass variance of the black and white pixels automatically. If the user is not satisfied with the value assigned by this automatic method, there is also provision for changing the threshold value using an additional control scheme. So, the final output $Y_f(t)$after automatic thresholding is

\begin{eqnarray}
I(i,j)&=&1 \mbox{ for } res^{j}_4(i)>th \\ \nonumber
&=&0 \mbox{ for } res^{j}_4(i)\leq th
\end{eqnarray}

\noindent where $th$ is the threshold value.

\begin{figure}[!h]
  \centering
  \includegraphics[width=5.2in,angle=0]{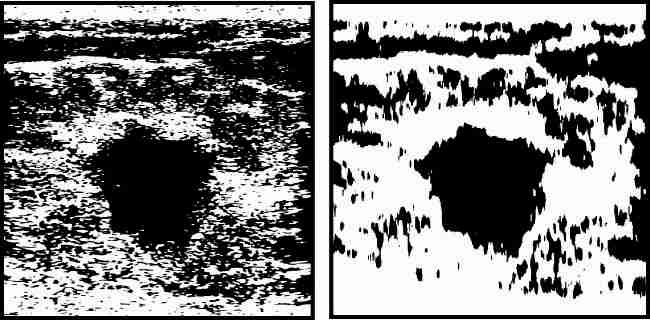}\\
  \caption{Result of automatic thresholding before and after empirical mode decomposition.}\label{Fig:7}
\end{figure}

\subsection{Initial lesion detection}

If multiple ROIs are identified through the thresholding segmentation, only one or two would be of diagnostic importance (belonging to abnormal lesions). To locate the abnormal lesions both the position and orientation (assuming 2D ultrasound images) have to be specified. We now manually choose the important ROI to identify the important ROI and analyze it. In future we opt to automatically select the desired ROI and thus imrove our desired segmentation approach.

\begin{figure}[!h]
  \centering
  \includegraphics[width=5.2in,angle=0]{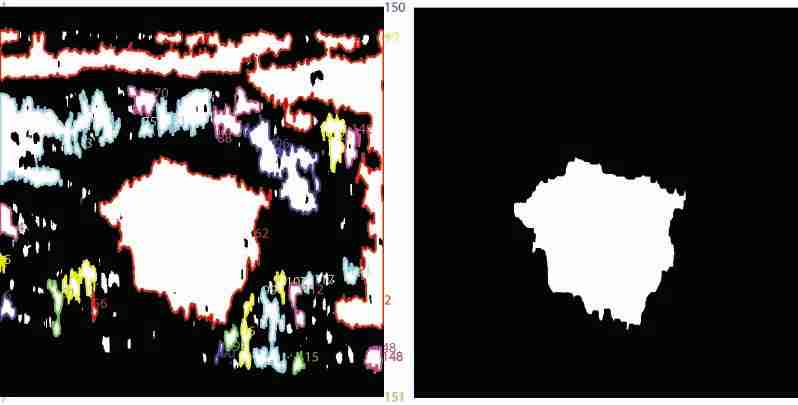}\\
  \caption{Boundary detection and ROI mask creation.}\label{Fig:8}
\end{figure}

\newpage
\chapter{Breast Tissue Characterization}
%Several sonographic and elastographic parameters can be used for robust characterization of breast lesions. Limitation of spectral and morphometric features in some cases enhanced the need of a new imaging modality called elastography which provides us with parameters like strain ratio for better categorization.

Several sonographic parameters can be used for robust characterization of breast lesions. These chapter includes the data acquisition process, the definition of the features and the method of obtaining these features \cite{kalam_main}.

\section{Materials and Methods}
From the acquired data, first an appropriate B-mode frame was selected by consulting with the radiologist and referring to the ultrasonic video file. Then the corresponding RF data of the frame was used for calculating the spectral parameters. The morphometric features were determined directly from the B-mode frame. The scanned frames were divided into several analysis regions. These regions called traces were identified separately with respect to the lesion such as left-anterior, tumor-anterior, right-anterior, left-lateral, tumor, right-lateral, left-posterior, tumor-posterior, and right-posterior. Only the tumor trace of the lesion was required for the majority of the quantitative features calculated. But posterior regions (tumor, left, and right) were needed for shadow measurements and the anterior regions were required for computing relative absorption.

Calibrated spectrum-analysis parameters were determined for calculating quantitative spectral features \cite{lizzi1,lizzi2,lizzi3}. Several steps were followed for calibrated spectrum analysis. A Hamming window was applied to the RF data; the windowed RF data went through a fourier transform, the resultant power spectrum determined and was expressed in dB. System and diffraction effects were then removed. Along with tissue properties, measured spectra depend on the combined two-way transfer function of the transducer and the ultrasonic system electronic modules, the two-way range-dependent diffraction function, and acoustic attenuation.

By scanning a uniform phantom, RF data was acquired and the electronic transfer function was estimated. The diffraction function was estimated using the same phantom. Spectral parameters were determined by subtracting the contribution from transfer function and diffraction. Then an empirical attenuation coefficient was used for attenuation correction. Finally, the linear-regression techniques were applied on the spectrum over the $6dB$ signal bandwidth. The slope of the regression line ($S$), its value at midpoint (midband fit) of signal bandwidth ($M$), and its intercept at zero frequency ($I$) are three important parameters. By sliding the Hamming window over all RF data and repeating the procedure, the images of these parameters were formed. The linear-regression line approximating the normalized power spectrum can be expressed as

\begin{eqnarray}
P(f)= I+sf \nonumber
\end{eqnarray}

\noindent where $I$ is spectral intercept, $s$ is slope, and $f$ is frequency. The midband fit,

\begin{eqnarray}
M = I + sf_0 \nonumber
\end{eqnarray}

\noindent where $f_0$ is the center frequency of the usable bandwidth. In the presence of linearly frequency-dependent attenuation, the linear regression line through the power spectrum can be expressed as,

\begin{eqnarray}
P_\alpha(f)=P(f) - 2\alpha df = I + (s-2\alpha d)f \nonumber
\end{eqnarray}

\noindent where  $\alpha$ is the effective attenuation coefficient and d is the depth of intervening tissue. \\

\noindent Thus, spectral intercept, $I_\alpha  = I$, \\
midband fit, $M_\alpha  = I  + (s - 2\alpha d)f_0$, \\
and slope, $s_\alpha  = (s - 2\alpha d)$. \\

\noindent Attenuation affects slope and midband fit, but intercept is not affected by attenuation. It is assumed that attenuation (in dB) varies linearly with frequency. The invariance of intercept in the presence of tissue attenuation has been proved to be true. In our work, midband fit and slope were corrected using an empirical value of 1.0 dB/MHz-cm \cite{mast}. Our spectrum analysis maintained the following specifications: window length, W = 2.4 mm, spectral bandwidth, B = 4 MHz (8-12MHz), and attenuation coefficient, = 1 dB/MHz-cm.

\begin{figure}[!h]
  \centering
  \includegraphics[width=5.5in,angle=0]{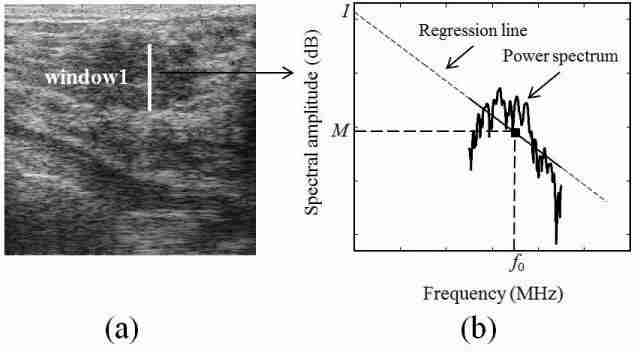}\\
  \caption{(a) A b-mode image and one window, (b) The power spectram of the corresponding window.}\label{Fig:80}
\end{figure}

\section{Characterization Features}
\subsection{Echogenicity}
Echogenicity is defined as the mean spectral intercept, within the lesion \cite{kalam_main}. No attenuation correction is necessary as spectral intercept is largely independent of frequency-dependent attenuation in the intervening media. The quantitative value of echogenicity is the mean value of $I$ within the lesion. Fibroadenomas were found to be more echogenic compared to the malignant tumors.

\subsection{Heterogeneity}
Heterogeneity can be defined in several ways. It is sometimes defined as the standard deviation of midband fit values, $\sigma_M$ within the lesion \cite{kalam_main}. The midband fit is equal to the average value of the spectrum over the measurement bandwidth. So, the midband fit is directly related to a widely employed parameter termed integrated backscatter or IB \cite{donell}. Heterogeneity of the lesion can be calculated by comparing   with  $M$ for a homogeneous region. As $M$ typically provides less noisy estimates compared to $I$ and $S$ permitting smaller departures from homogeneity to be detected, we selected $\sigma_M$ to provide an index of tissue heterogeneity. The fibroadenomas were found to be more homogenous compared to the malignant tumors.

Heterogeneity also may depend on texture.  $M$ contains no textural information. So, texture of midband fit inside the lesion was also computed, defined in terms of a four-neighborhood pixel algorithm (FNPA) \cite{yao} and Hurst Coefficient fractal dimension measure \cite{hurst}. Among the 4-neighborhood-pixels algorithm (FNPA) and the histogram algorithm (HA), defined by Yao et al. \cite{yao} one of the FNPAs is a good descriptor of texture, defined for an image of size $m X n$ with pixel values $x(k,l)$ as

\begin{eqnarray}
FP_2=FP_1/\mu   \nonumber
\end{eqnarray}

\noindent where,

\begin{eqnarray}
FP_1 = \sum_{l=1}^{n} \sum_{k=1}^{m} \frac{1}{4} &[& |x(k,l)-x(k-1,l)| + |x(k,l)-x(k,l-1)| \\ \nonumber
&+& |x(k,l)-x(k+1,l)| + |x(k,l)-x(k,l+1)|]  \nonumber
\end{eqnarray}

\noindent $\mu$ being the mean value of $FP_1$. Our implementation varies slightly from Yao et al, in which a linear regression parameter was subtracted from $FP_1$. This normalization is not required as transmit power is constant in all scans and as we compensate for attenuation ($\alpha$).

	We also implemented the co-occurrence matrix to estimate B-mode texture \cite{garra}. Chen et al. \cite{chen} used texture correlation between neighboring pixels within sonographic images for classifying breast tumors. Because the normalized autocorrelation coefficients. can reflect the inter-pixel correlation within a digital image, these coefficients can be used as the texture features of a tumor. In general, the two-dimensional (2-D) autocorrelation coefficients are further modified into a mean-removed version to have the similar autocorrelation features for the images with different brightness but with a similar texture. That is, the 2-D normalized autocorrelation coefficient between pixel $(i,j)$ and pixel $(i+\triangle m,j+\triangle n)$ in an image $f$, with size $M x N$ is defined as:

\begin{eqnarray}
\gamma(\triangle m, \triangle n) = \frac{A(\triangle m, \triangle n)}{A(0,0)}, \nonumber
\end{eqnarray}

where

\begin{eqnarray}
A(\triangle m, \triangle n) &=& \frac{1}{(M-\triangle m)(N- \triangle n)} \sum_{x=0}^{M-1-\triangle m} \sum_{y=0}^{N-1-\triangle n}|(f(x,y)-f') \\ \nonumber
&.&(f(x+\triangle m, y+ \triangle n)-f')|   \nonumber
\end{eqnarray}

$f'$  is the mean value of $f(x,y)$.

\subsection{Morphometric features}	
The aspect ratio is tumor depth divided by width. In larger carcinomas, this criterion is less useful due to their more irregular shapes and growth along duct axes. We define Aspect Ratio as the maximum vertical lesion-dimension divided by maximum horizontal lesion-dimension. The aspect ratio is found to be smaller for benign cases compared to the malignant ones. \\

\begin{eqnarray}
Aspect ratio = \frac{Tumor \ depth}{Tumor \ width} \nonumber
\end{eqnarray}

\begin{figure}[!h]
  \centering
  \includegraphics[width=6in,angle=0]{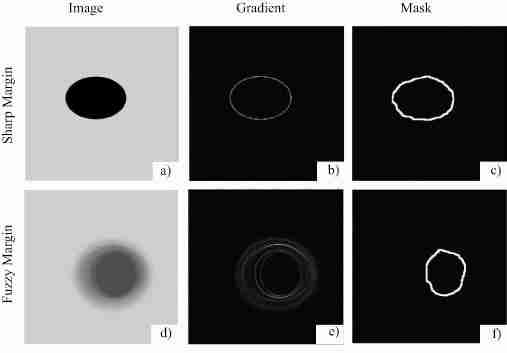}\\
  \caption{(a),(d): Original images, (b),(e): Gradient images, (c),(f): Mask for margin definition calculation.}\label{Fig:mdef}
\end{figure}

Invasive ductal carcinomas generally have fuzzy borders as a result of having invading margins. On the contrary, cancers that elicit little desmoplastic reaction (proliferation of fibroblasts) typically have clear margins, but are highly irregular in shape. We define Margin Definition as the sum of the magnitude of the gradient of $M$ on a lesion contour normalized by the sum of magnitude of $M$ on the lesion contour \cite{kalam_main}. Although this feature uses both the lesion contour as well as a spectral parameter, we used this as a spectral feature. As $M$ is statistically well-behaved, is relatively speckle-free, and can more easily be corrected for system effects and diffraction, the midband fit image was used instead of the B-mode (envelope of RF echoes) image. Benign lesions exhibit greater value of gradient-based margin definition. Figure \ref{Fig:mdef} shows two lesion images with sharp \& fuzzy borders and shows their gradient images that are used for calculating margin definition. The lesion with sharp border gives the value of $0.2898$ and the lesion with fuzzy border gives the value of $0.0265$.\\

A convexity parameter is the ratio between convex perimeter and actual lesion perimeter; this parameter can be used to express border irregularity; this also is an excellent descriptor of spiculation. The fractal dimension is found to be lower for benign lesions and higher for malignant ones whereas the convexity is higher for benign cases and lower for the malignant tumors \cite{kalam_main}. \\
	
\begin{eqnarray}
Convexity = \frac{Convex \ perimeter}{perimeter} \nonumber
\end{eqnarray}

\begin{eqnarray}
Solidity = \frac{Area}{Convex \ area} \nonumber
\end{eqnarray}

\begin{figure}[!h]
  \centering
  \includegraphics[width=5.5in,angle=0]{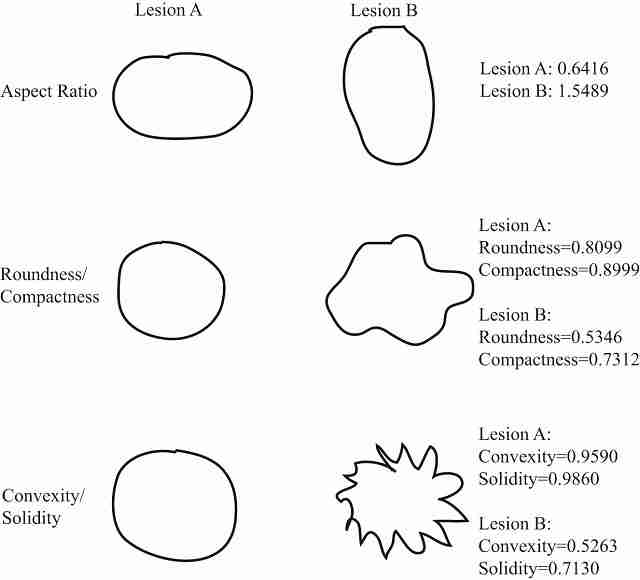}\\
  \caption{Different lesion shapes and their morphological features.}\label{Fig:shape}
\end{figure}

The Compactness is defined as the ratio of square root of the surface area to the maximum diameter; therefore it is sensitive to shape of the lesion.  Roundness is the ratio of area and maximum diameter squared. Compactness is the squareroot of roundness. Formfactor is the ratio of area and perimeter squared \cite{kalam_main}. Figure \ref{Fig:shape} shows the values of some morphological features for different lesion shapes.

\begin{eqnarray}
Compactness = \frac{\sqrt{Tumor \ surface \ area}}{Maximum \ diameter} \nonumber
\end{eqnarray}

\begin{eqnarray}
Roundness = \frac{Tumor \ surface \ area}{Maximum \ diameter^2} \nonumber
\end{eqnarray}

\newpage
\chapter{Results}

To determine the clinical value of the developed algorithms, their effectiveness was tested on data collected from patients with known histopathology of the breast lesions. The IRB (Institutional Review Board) approval was received for this study. Informed consent was collected from each subject. Bangladesh University of Engineering and Technology (BUET) medical center was used as the data collection center. A SonixTOUCH Research ultrasound machine with L14-5/38 linear transducer was used in scanning the patients. More than one expert radiologists/sonologists actively participated in the data collection process. These data was sampled at 40 MHz and analyzed in our research lab using various algorithms and the lesions were classified as benign or malignant lesions on the basis of our ultrasonic quantitative parameters derived from B-mode image and RF data.

\newpage

\begin{figure}[!h]
  \centering
  \includegraphics[width=4.5in,angle=0]{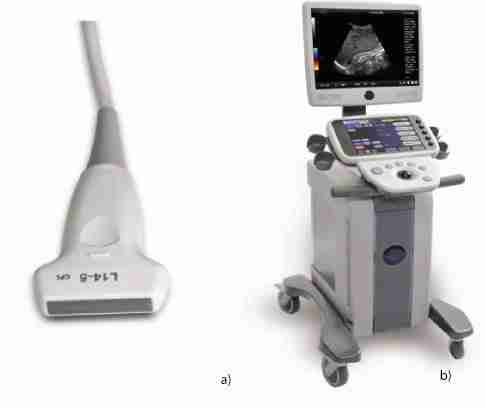}\\
  \caption{(a) The L 14-5/38 Linear Transducer, (b) The sonixTOUCH research machine.}\label{Fig:80}
\end{figure}

\begin{figure}[!h]
  \centering
  \includegraphics[width=5.5in,angle=0]{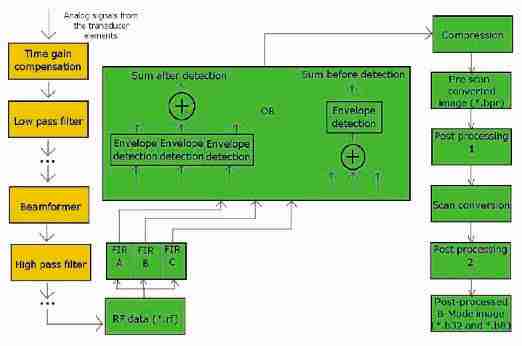}\\
  \caption{Processing Pipeline of the Machine.}\label{Fig:80}
\end{figure}

\section{Objective Diagnosis}

\subsection{Semi-automatic segmentation}
The analysis was done on $64$ lesions with a mean dimension of $12.0391$mm. The extreme points of the lesions in the b-mode image were already marked by the sonologists and the dimension of the ROIs were compared with it. The mean error was found to be $0.1305$mm. The percentage error was $2.80772 \%$.

\begin{figure}[!h]
  \centering
  \includegraphics[width=5.5in,angle=0]{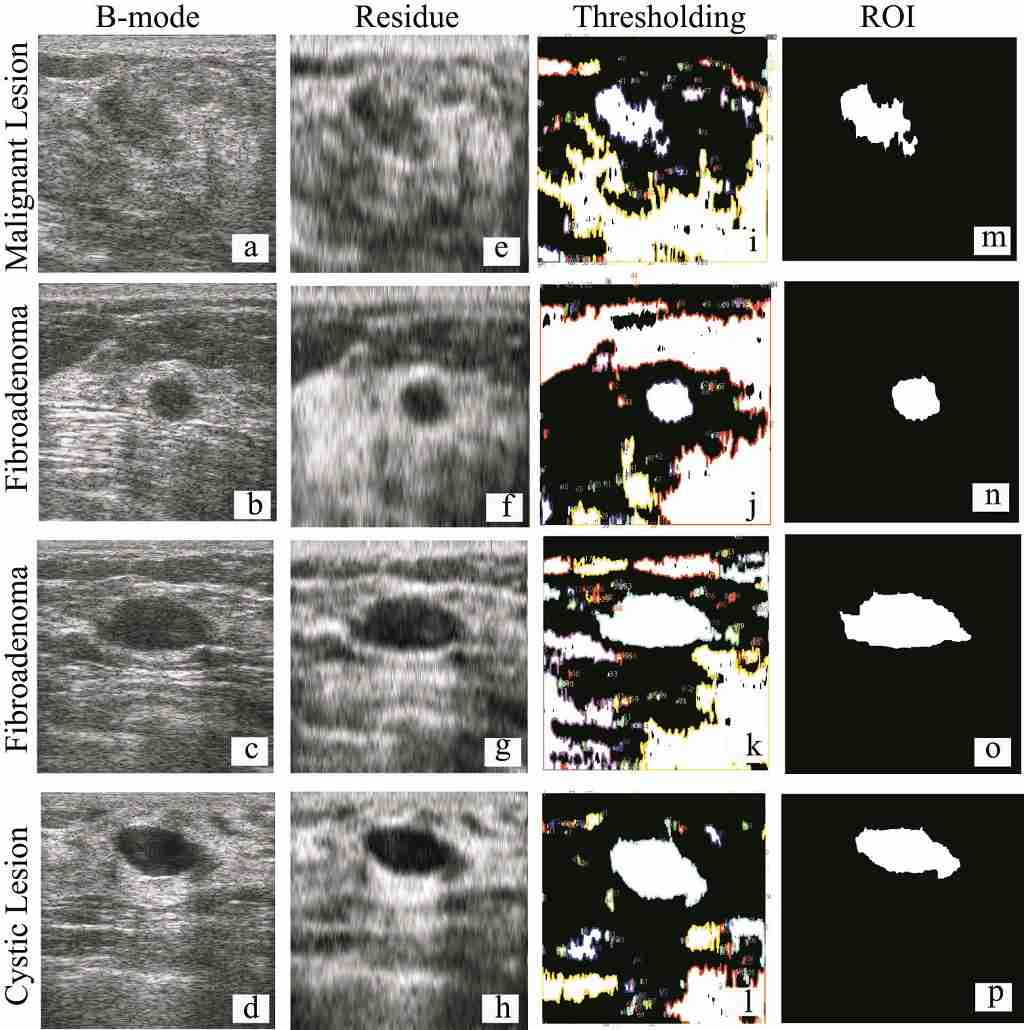}\\
  \caption{Lesion segmentation for different types of lesions. (a)-(d) are b-mode images generated from RF data,
  (e)-(h) are residues after EMD, (i)-(l) are the images after thresholding, (m)-(p) are the final ROIs.}\label{Fig:seg_big}
\end{figure}

\begin{figure}[!h]
  \centering
  \includegraphics[width=6.5in,angle=0]{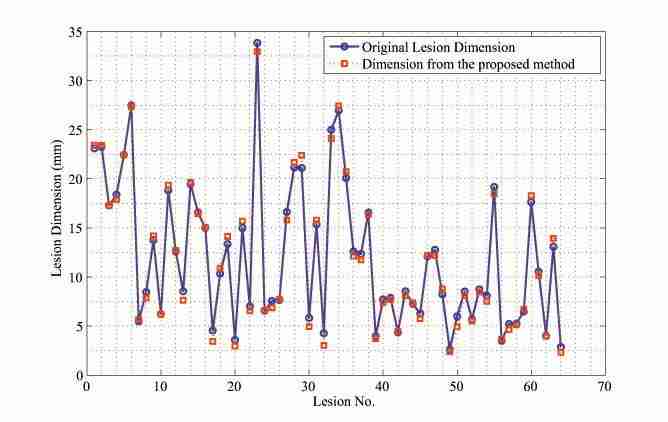}\\
  \caption{Comparative scenario of original lesion dimension and dimension from the proposed method.}\label{Fig:seg_com}
\end{figure}

In figure \ref{Fig:seg_big}, segmentation for different types of lesion has been presented. (a)-(d) are b-mode images generated from RF data,
  (e)-(h) are residues after EMD, (i)-(l) are the images after thresholding and (m)-(p) are the final ROIs. Figure \ref{Fig:seg_com} shows comparative scenario of original lesion dimension and dimension from the proposed method.

\newpage

\newpage

\subsection{Breast tissue characterization}
The spectral, morphometric and elastographic features were  extracted from our own database. The algorithms were implemented on 116 patients. The results are given in tables \ref{Table:t1}-\ref{Table:t2}.

\begin{table}[!h]
  \centering
  \caption{Extracted values of spectral features}\label{Table:t1}
  \begin{tabular}{|l|l|l|}
    \hline
    \textbf{Feature} & \textbf{Benign} & \textbf{Malignant}\\ \hline \hline
    Echogenicity (dB) & $3.1884 \pm 8.2389$ & $-2.8200 \pm 9.6313$\\ \hline
    Heterogeneity (dB) & $7.3728 \pm 2.3343$ & $8.9937 \pm 2.8422$\\ \hline
    FNPA(dB) & $0.1551 \pm 0.1910$ & $0.2040 \pm 0.1803$\\ \hline
    FNPA & $0.4879 \pm 0.0853$ & $0.4621 \pm 0.0814$\\ \hline
    Cooccurance Contrast  & $12.5541 \pm 5.0514$ & $9.8357 \pm 3.1157$\\ \hline
    Hurst Coefficient (dB) & $0.5347 \pm 0.0945$ & $0.5160 \pm 0.0624$\\ \hline
    Hurst Coefficient & $1.0269 \pm 0.4311$ & $0.8768 \pm 0.1789$\\ \hline
    Margin Definition & $0.1546 \pm 0.0672$ & $0.1470 \pm 0.0692$\\ \hline
  \end{tabular}
\end{table}

\begin{table}[!h]
  \centering
  \caption{Extracted values of morphological features}\label{Table:t2}
  \begin{tabular}{|l|l|l|}
    \hline
    \textbf{Feature} & \textbf{Benign} & \textbf{Malignant}\\ \hline \hline
    Aspect Ratio & $0.6945 \pm 0.2225$ & $0.9883 \pm 0.3997$\\ \hline
    Compactness & $0.7110 \pm 0.0933$ & $0.7285 \pm 0.0701$\\ \hline
    Roundness & $0.5141 \pm 0.1336$ & $0.5353 \pm 0.0997$\\ \hline
    Convexity & $0.8303 \pm 0.0325$ & $0.8180 \pm 0.0346$\\ \hline
    Solidity  & $0.9160 \pm 0.0436$ & $0.9026 \pm 0.0536$\\ \hline
%    Formfactor & $0.6763 \pm 0.1158$ & $0.6945 \pm 0.1165$\\ \hline
  \end{tabular}
\end{table}

%\section{Subjective Diagnosis}
%
%\begin{table}[!h]
%  \centering
%  \caption{Subjective Diagnosis}\label{Table:FontSizes}
%  \begin{tabular}{|l|l|l|}
%    \hline
%    % after \\: \hline or \cline{col1-col2} \cline{col3-col4} ...
%    \textbf{} & \textbf{US} & \textbf{US+ES}\\ \hline \hline
%    Sensitivity & $40\%$ & $100\%$\\ \hline
%    Specificity & $98.51\%$ & $93.59\%$\\ \hline
%    Positive Predictive Value & $85.71\%$ & $80\%$\\ \hline
%    Accuracy & $87.80\%$ & $94.90\%$\\ \hline
%
%  \end{tabular}
%\end{table}

\section{Correlation with Histopathologic Findings}
\begin{figure}[!t]
  \centering
  \includegraphics[width=6in,angle=0]{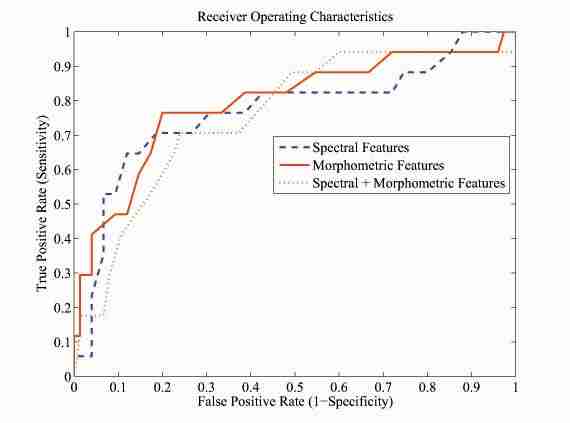}\\
  \caption{Receiver operating characteristics of spectral, morphometric and combined features.}\label{Fig:roc}
\end{figure}

Among the quantitative features, echogenicity, heterogeneity, margin definition, aspect ratio and convexity were finally used for characterization as their values show higher separation between benign and malignant class. After being given optimum weights, the combination of the featuress gave highest separation \& the spectral and morphological features were first used separately and then used together to charaterize the lesions. The results are summerized in table \ref{Table:corr}. For characterization using spectral features, echogenicity, heterogeneity and margin definition were given weights of 0.5, 0.25 and 0.25 respectively. For characterization using morphometric features, aspect ratio and convexity were given weights of 0.5, and 0.5. And for combined characterization, echogenicity, heterogeneity, margin definition, aspect ratio and convexity were given weights of 0.14, 0.14, 0.07, 0.36 and 0.29 respectively. As expected, the sensitivity, specificity and the area under ROC curve increased in the combined case than each of the individual one. The corresponding ROC curves are shown in figure \ref{Fig:roc}.

\begin{table}[!h]
  \centering
  \caption{Correlation with histopathologic findings}\label{Table:corr}
  \begin{tabular}{|l|l|l|l|}
    \hline
    \textbf{} & \textbf{Spectral} & \textbf{Morphometric} & \textbf{Spectral + Morphometric}\\
    \textbf{} & \textbf{Features} & \textbf{Features} & \textbf{Features}\\ \hline \hline
    Sensitivity & $70.588\%$ & $70.588\%$ & $76.471\%$\\ \hline
    Specificity & $73.333\%$ & $76\%$ & $80\%$\\ \hline
    Area Under ROC Curve & $76.902\%$ & $75.608\%$ & $79.294\%$\\ \hline

  \end{tabular}
\end{table}

\newpage
\chapter{Conclusion}

\section{Summary}
Automatic segmentation of ultrasound images is essential for robust characterization of breast tissues. So, before characterizing the breast lesions, we developed a semi-automatic segmentation algorithm.

In this thesis, we have presented a novel method for semi-automatic segmentation using empirical mode decomposition. This analysis was done on $64$ lesions with a mean dimension of $12.0391$mm. The mean error was found to be $0.1305$mm. The percentage error was $2.80772 \%$.

We have used ultrasonic features for differentiating between benign and malignant lesions. Among the quantitative features, echogenicity, heterogeneity, margin definition, aspect ratio and convexity were finally used. We used optimum weights for combining the features to achieve the highest possible separation. For characterization using spectral features; echogenicity, heterogeneity and margin definition were given weights of 0.5, 0.25 and 0.25 respectively which gave a sensitivity of $70.588\%$, specificity of $73.333\%$ and an area under ROC curve of $76.902\%$. For characterization using morphometric features; aspect ratio and convexity were given weights of 0.5 and 0.5 which gave a sensitivity of $70.588\%$, specificity of $76\%$ and an area under ROC curve of $75.608\%$. And for combined characterization; echogenicity, heterogeneity, margin definition, aspect ratio and convexity were given weights of 0.14, 0.14, 0.07, 0.36 and 0.29 respectively which gave a sensitivity of $76.471\%$, specificity of $80\%$ and an area under ROC curve of $79.294\%$.

\section{Limitations and Future Scope}
We did not compare our EMD based segmentation algorithm with other prevailing algorithms for segmentation. Our segmentation process is semi-automatic as it requires manual thresholding in some cases. Some of the characterization parameters such as margin definition could have been more decisive if some different algorithm was used for its extraction. However, we intend to include the comparison of our segmentation algorithm with other methods in our future work. We believe working in EMD domain will bring out overwhelming results is different aspects such as B-mode enhancement, extraction of features etc. We hope to make our segmentation algorithm fully automatic in future.

%We developed a 2-D plot using sonographic feature in one axis and elastographic features in another using 47 lesions. The plot shows accurate separation between fibroadenoma, cyst and malignant lesion. The elastographic feature is strain ratio and the sonographic feature is gray scale ratio (related to the amplitude of Rf data). We ae trying to develop a physical relation between this gray scale ratio and amplitude so that our plot becomes justified.

%\input{bib.tex}

\end{document}